\documentclass[conference]{IEEEtran}
\usepackage{blindtext, graphicx}
\usepackage{float}
\usepackage[section]{placeins}
\usepackage{array}
\usepackage{tabularx, booktabs}
\newcolumntype{P}[1]{>{\centering\arraybackslash}p{#1}}
\usepackage{authblk}
\usepackage{cite}
\usepackage{lettrine}
\usepackage{fancyhdr}
\usepackage{caption}
\ifCLASSINFOpdf
\else
\fi
\begin{document}
%
% paper title
% can use linebreaks \\ within to get better formatting as desired
\pagestyle{fancy}
\title{Deep Learning based Isolated Arabic Scene Character Recognition}

% author names and affiliations
% use a multiple column layout for up to three different
% affiliations
\author[1,3]{Saad Bin Ahmed}
\author[2]{Saeeda Naz}
\author[1]{Muhammad Imran Razzak}
\author[3]{Rubiyah Yousaf}
\affil[1]{King Saud bin Abdulaziz University for Health Sciences, Riyadh, 11481, Saudi Arabia
\authorcr Email: {\{ahmedsa, razzaki\}@ksau-hs.edu.sa}}
\affil[2]{GGPGC No.1, Abbottabad, Higher Education Department, Khyber Pakhtunkhua (KPK), Pakistan
\authorcr Email: { \{saeeda292\}@gmail.com}}
\affil[3]{Malaysia Japan Institute of Information Technology (MJIIT), Universiti Teknologi Malaysia, KualaLumpur, Malaysia
\authorcr Email: {\{rubiyah.kl\}@utm.my}\vspace{1.5ex}}

% conference papers do not typically use \thanks and this command
% is locked out in conference mode. If really needed, such as for
% the acknowledgment of grants, issue a \IEEEoverridecommandlockouts
% after \documentclass

% for over three affiliations, or if they all won't fit within the width
% of the page, use this alternative format:
% 
%\author{\IEEEauthorblockN{Michael Shell\IEEEauthorrefmark{1},
%Homer Simpson\IEEEauthorrefmark{2},
%James Kirk\IEEEauthorrefmark{3}, 
%Montgomery Scott\IEEEauthorrefmark{3} and
%Eldon Tyrell\IEEEauthorrefmark{4}}
%\IEEEauthorblockA{\IEEEauthorrefmark{1}School of Electrical and Computer Engineering\\
%Georgia Institute of Technology,
%Atlanta, Georgia 30332--0250\\ Email: see http://www.michaelshell.org/contact.html}
%\IEEEauthorblockA{\IEEEauthorrefmark{2}Twentieth Century Fox, Springfield, USA\\
%Email: homer@thesimpsons.com}
%\IEEEauthorblockA{\IEEEauthorrefmark{3}Starfleet Academy, San Francisco, California 96678-2391\\
%Telephone: (800) 555--1212, Fax: (888) 555--1212}
%\IEEEauthorblockA{\IEEEauthorrefmark{4}Tyrell Inc., 123 Replicant Street, Los Angeles, California 90210--4321}}

% use for special paper notices
%\IEEEspecialpapernotice{(Invited Paper)}

% make the title area
\maketitle

\begin{abstract}
%\boldmath

The technological advancement and sophistication in cameras and gadgets prompt researchers to have focus on image analysis and text understanding. The deep learning techniques demonstrated well to assess the potential for classifying text from natural scene images as reported in recent years. There are variety of deep learning approaches that prospects the detection and recognition of text, effectively from images. In this work, we presented Arabic scene text recognition using Convolutional Neural Networks (ConvNets) as a deep learning classifier. As the scene text data is slanted and skewed, thus to deal with maximum variations, we employ five orientations with respect to single occurrence of a character.
The training is formulated by keeping filter size $3$ x $3$ and $5$ x $5$ with stride value as $1$ and $2$.
During text classification phase, we trained network with distinct learning rates.
Our approach reported encouraging results on recognition of Arabic characters from segmented Arabic scene images.

\end{abstract}
% IEEEtran.cls defaults to using nonbold math in the Abstract.
% This preserves the distinction between vectors and scalars. However,
% if the journal you are submitting to favors bold math in the abstract,
% then you can use LaTeX's standard command \boldmath at the very start
% of the abstract to achieve this. Many IEEE journals frown on math
% in the abstract anyway.

% Note that keywords are not normally used for peerreview papers.

% For peer review papers, you can put extra information on the cover
% page as needed:
% \ifCLASSOPTIONpeerreview
% \begin{center} \bfseries EDICS Category: 3-BBND \end{center}
% \fi
%
% For peerreview papers, this IEEEtran command inserts a page break and
% creates the second title. It will be ignored for other modes.
\IEEEpeerreviewmaketitle

\section{Introduction}
\lettrine[lines=2]{\bf{T}}{he} content based image analysis has obtained popularity in recent years. 
The most complex part of content based image analysis is scene text recognition which is categorized as a special problem in the field of  Optical Character Recognition (OCR). 
In OCR, the techniques and methods that have applied on cleaned machine rendered and synthetic images produced desired results. 
It is considered as a solved problem for most of the scripts. 
However, due to infancy of scene text recognition, it is struggling towards accuracy particularly in cursive scripts~\cite{str1, str2, str3, str4, str5}.

The scene image having a text captured from camera has built-in complex noise associated to it.
The detection and recognition of text from scene text images is considered as subtle task because there may be non-text elements in an image which should be detected and removed before applying any classification technique on such intrinsic images. 
We can not process scene text data or printed data in a same way. 
The techniques and methods that have already been applied on printed and clean scanned data drastically failed on recognition of scene text data. 
Because captured images do not have only textual information, instead we need to tackle non-text objects and the issue of text appearance in various colors, formats, and sizes that make it harder to apply automated tools to detect and eliminate such irrelevant data.

The probable applications of scene text recognition is to assist the visually impaired, number plate recognition, intelligent vehicle driving systems, machine language translation and may provide help in machine reading for robotics systems.
The image content depicts the intuitive information, each with a different challenge. 
Among various challenges the most prominent is orientation and size of a text in a scene image. 

The scene text recognition has been divided into three phases by most of the authors ~\cite{it,td,tr,tm,icdar2011,ah, es}.  
These stages included text segmentation or localization, text extraction and text recognition. 
In every phase, intense preprocessing is required to accomplish the task~\cite{it}. 
In text segmentation or localization, we detect text area in presence of other objects in an image, while extraction means to segment the text carefully so that it may recognized in last stage by OCR technique. 
It is obvious that OCR will not directly process the video image because as mentioned before the nature of OCR is more towards to process clean document images taken in standard resolution and in specific settings. 
The video images often has color blending, blur, low resolution and complicated background in presence of different objects. 
It is hereby assumed that scene text and video text shares same sort of problems and difficulties in the recognition. 

Most of scene text recognition techniques have been witnessed on Latin or English text. 
The  database plays a vital role in evaluation of state-of-the-art techniques. 
Some scene text datasets are available for Latin script~\cite{it,tm}. 
The cursive script is not thoroughly investigated by researchers yet. 
The availability of benchmark or large size dataset is a fundamental requirement for training and testing the state-of-the-art classifiers in scene text recognition. 
Therefore, the acquisition of scene text images, development of scene text based database, and its distribution to the researchers for comparison of different techniques and methods is one main focus of attention.
We have prepared and compiled  Arabic scene text data and consider its subset for evaluation on Convolutional Neural Network (ConvNet).

In this paper, we evaluate the potential of ConvNets on Arabic scene text recognition.
The Arabic scene was segmented from captured images.
The preprocessing was performed for uniform representation of segmented data before passed them to classifier.
We performed experiments on different parameters variations that reveals satisfactory results.

The rest of this paper includes related work as presented in Section II.
The proposed methodology including feature extraction technique and description about learning classifier and dataset is elaborated in Section III whereas in Section IV, we managed to explain about our experimental parameters and their settings.
This section further discuss about learning accuracy and influential parameters.
Section V summarized our work under conclusion.

\section{Related Work}
In scene text recognition, text detection and segmentation pose a great challenge. 
Once text have been segmented correctly then there is a need to extract features from segmented text image and pass it to the classifier, this is how machine learning approaches work. 
We have compiled few latest work presented, so that we may know about how much work has been done in this field by keeping in view the Arabic or cursive script.
The efficient scene text localization and recognition technique is proposed by~\cite{es}.
They used region based text  detection which refine text hypothesis with the assumption that all characters are spotted through connected component. Their proposed technique executed in real time and have been evaluated on ICDAR 2013 dataset. A complete system for text detection and localization in gray scale images is proposed by~\cite{td}.  
The boosting framework integration feature in combination to the computational complexity approach named weak classifier is developed to the make efficient text detector. 
They evaluate their proposed scheme on ICDAR 2003 robust reading and text localizing dataset. 
Their proposed technique performed well on various font sizes, styles, and types exist in natural scenes. Another approach by [8] proposed text localization using conditional random fields. The preprocessing is performed by conversion of color image into grayscale and then make histogram of oriented gradients as a feature. The connected component analysis was performed after analysis of text and non-text regions by conditional random fields. Their proposed technique gives better results in comparison to ICDAR 2003 competition dataset. 

The color based approach for text detection of Farsi text is proposed by~\cite{ah}. 
The text images are then detected by fusion of color and edge information. The extracted text are verified by wavelet histogram and histogram of oriented gradient. They reported effective results on their large dataset.
The work on Arabic text extraction from video images is proposed by~\cite{at}. 
They used synthetic text images taken from numerous news channels. 
They localize and segment the Arabic text encrypted in video. The text and background pixels were determined through thresholding that produced binary image. They also maintained the temporal information of a video image for verification purpose. They reported their experimentation results on their own proposed dataset as robust.
In recognition phase of scene text images, OCR techniques applies for learning of a text and recognition purpose. The evaluation of cursive and non-cursive scripts using Recurrent Neural Network is proposed by~\cite{ec}. The cursive script's experiments were performed on large Arabic script synthetic dataset. They reported encouraging results on both scripts. Another effort to develop a standard handwritten Arabic Nastal'iq script is compiled by~\cite{ucom}.
They gathered handwritten text from 500 individuals which is evaluated by Bidirectional Long Short Term Memory network~\cite{hu}. However, we use Bidirectional Long Short Term  Memory (BLSTM) network as a classifier to learn the detected scene images.

There exist some algorithms using Scale-Invariant Feature Transform (SIFT).~\cite{sift} proposed a very interesting technique for scene text recognition using SIFT vector. They proposed novel approach for Scale based region growing algorithm.  They used SIFT keypoints to manipulate the local text region. The SIFT algorithm known as an efficient technique. By using it in their proposed work the keypoint  extraction time drops down exponentially in comparison to~\cite{at}. They evaluate their technique on two publicly available datasets i.e., MSRG and ICDAR. They reported good results on their proposed algorithm in comparison to~\cite{cvpr} and~\cite{icip} on same datasets. 
The multi frame scene text recognition in video images is presented by~\cite{icme}. 
They developed a framework on Scene Text Character (STC) recognition for predicting the character and conditional random field was used for word spotting. The STC features were taken from SIFT descriptors and Fisher vector. 
They also collected the dataset from natural scene videos and extract text from it. They evaluated their algorithm on their own collected dataset and three bechmark datasets i.e., CHARS74K, ICDAR2003, ICDAR2011 as reported in their manuscript. The results were conducted on single frame and multiframe scene text and conclude that their approach performs much better on multiframe scene text. 
	
All presented state-of-the-art techniques have been evaluated on Latin and Chinese script but Arabic script is not been addressed in more detail by these mentioned techniques. The availability of dataset is essential for the purpose to assess the performance of proposed algorithm.
By keeping this forefront, we presented state-of-the art based approach on Arabic scene text recognition.

\section{Methodology}
In this manuscript we proposed ConvNets for Arabic scene text recognition. 
ConvNets is type of deep learning Neural Network that is based on the idea of multilayer perceptrons (MLPs).
It has been successfully applied on recognition of various objects in image.
Unlike Recurrent Neural Networks (RNNs), ConvNets is more focused on single instance learner rather a sequence learner. 
The context is not important for ConvNets training.
Nowadays, ConvNets are considered as an important tool in machine learning applications~\cite{str3, str4, str5}.
The Arabic script is complex and cursive in nature.
Various authors have reported work on synthetic and scanned Arabic text but very few works are presented on Arabic scene text recognition till date.
\begin{figure*}[h]
\centering
\includegraphics[scale=0.7]{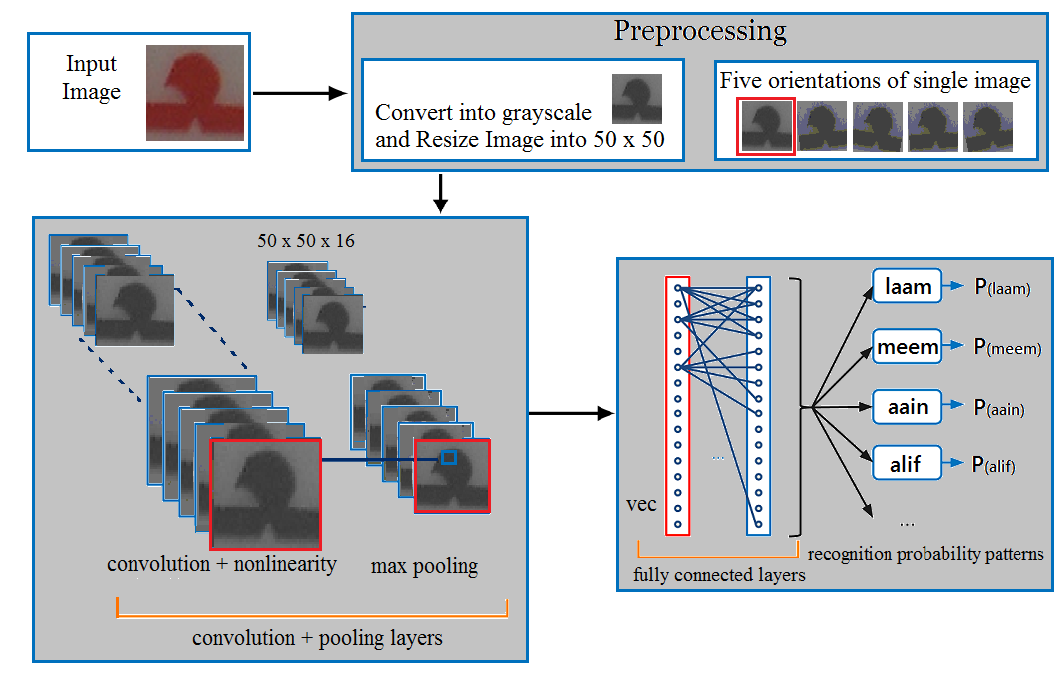}
\caption{\bf Proposed methodology based on ConvNets}
\label{methodology}
\end{figure*}	
In Figure~\ref{methodology}, the input image of arbitrary size is preprocessed with respect to size ($50$ x $50$) and converted it into gray scale. 
The image is saved with five various orientations. 
With oriented images we are processing five images against one input image.
The convolution is performed and features were extracted from pooling.
The detail about feature extraction is mentioned in the following sub-section.
In the last stage fully connected layers classified the given image and compute the probability by keeping in view the current input image.

\subsection{ConvNets as a Feature Extractor}

Suppose, we have relatively big image in size and we want to extract and learn $70$ features from each image. 
The architecture we used is fully connected feed forward network.
In this situation, the computation would be so complex and takes much time to process a single epoch. 
Even in backpropagation the computation would be slower.

By keeping in mind the performance measure, In ConvNets, the solution is to limit the connections between hidden units and input units.
By this, hidden unit will connect only a subset of input units. 
In particular, each hidden unit will connect to small group of contagiously located pixels in input unit.
The image volume $I_v$ is computed by width $w$, height $h$ and depth $d$.
\begin{equation}
    I_v = w+h+d
\end{equation}

Lets assume, number of filters as $k$, 
the spatial extent as $f$, the stride as $s$, and amount of zero padding $p$.
Here the zero padding is relevant to linear output. The non linear output is represented as a negative values which is replaced by zero to get linear layer output.
At each location where filter process and moving as stride dictates, the $w$ and $h$ is computed for each kernel as follows,
where $W_i$ and $H_i$ are width and height of $ith$ kernel.
The number of kernels make the depth $d$.
\begin{equation}
W_i= (w_1-f+2p)/s+1
\end{equation}

\begin{equation}
H_i= (h_1-f+2p)/s+1
\end{equation}

\begin{figure}[H]
\centering
\includegraphics[scale=0.8]{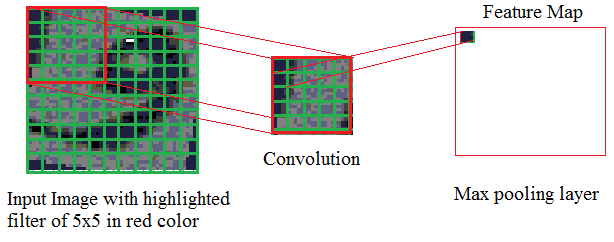}
\caption{\bf Feature extraction using ConvNets}
\label{feature}
\end{figure}

    As shown in Figure~\ref{feature}, the filter was sliding over the whole image. 
    At each time when it stops (dictated by stride), it takes a maximum value as a feature from involved pixels and write at (1,1) of output layer. 
    When stride value is $1$, it means the filter will move one pixel to the right and will perform the same operation as previously mentioned.
    After performing operation in one row, it will move one down and begin the entire process again until it process whole image.

\subsection{ConvNets as a Learning Classifier}
Although ConvNets is suitable for feature extraction but it can be used as a learning classifier. 
In our proposed work we used ConvNets as our classification technique. 
We used fully connected $3$ x $3$ and $5$ x $5$ spatial convolution kernels architecture with max pooling strategy as represented in equation,

\begin{equation}
F^{`}(x)= max_k f(x_{sj})
\end{equation}

The max pooling strategy takes maximum value $max_k$ from the filter which is been observed on pixel $x_{sj}$. 
The Rectified Linear Unit (ReLU) is used as an activation function which removes the non-linearity of processed data.
The features that have been learned through training is compared with extracted features of testset data.
The difference is computed and accuracy is measured. 
The output neurons in the proposed network are represented as activation of each class. 
The most active neuron analogously predict the class for given input. 
The softmax layer is used to interpret the prediction about activation value for each class.

\section{Results and Discussions}
The details about dataset and performed experiments are mentioned in the following sub-sections.

\subsection{Dataset}
\begin{figure}[H]
\centering
\includegraphics[scale=0.3]{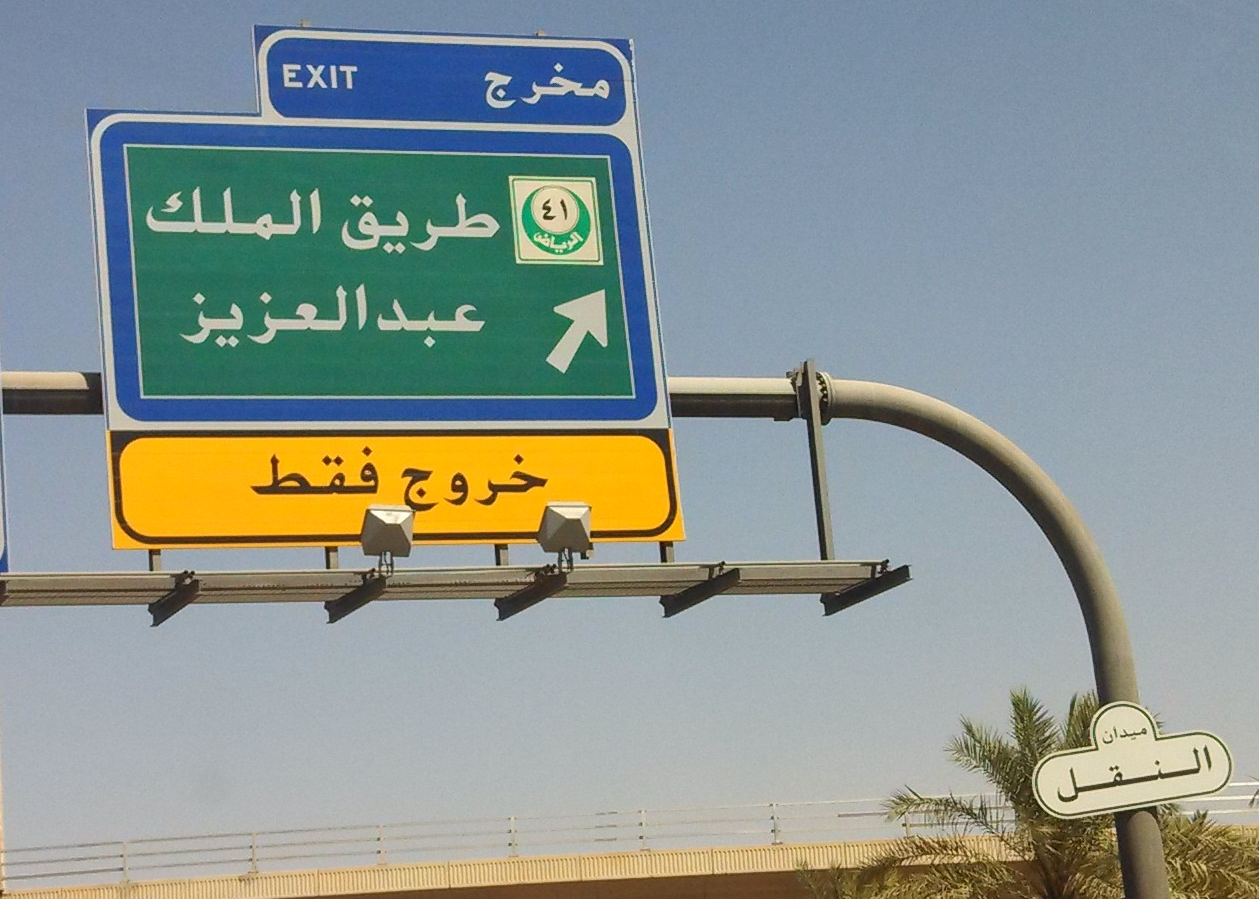}
\caption{\bf Sample Arabic scene text image}
\label{sample}
\end{figure}

We have extracted Arabic images from EAST (English-Arabic Scene Text) dataset. 
The Arabic scene text sample is presented in Figure~\ref{sample}.

\begin{figure}[H]
\centering
\includegraphics[scale=0.6]{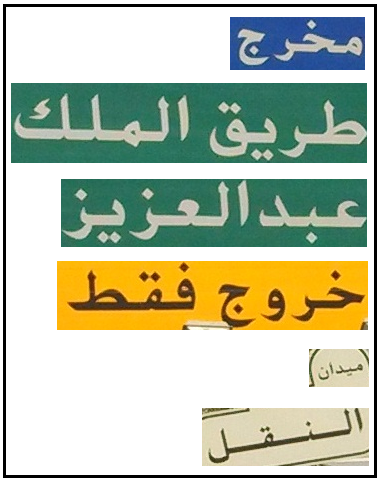}
\caption{\bf Segmented Arabic text lines from natural image}
\label{segmented}
\end{figure}

\begin{figure}[H]
\centering
\includegraphics[scale=0.9]{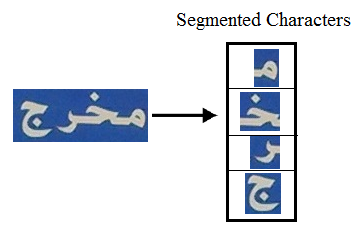}
\caption{\bf Character segmentation of a word}
\label{char}
\end{figure}

In Arabic, it is cumbersome to disintegrate the word into individual characters because of different shape variations with respect to character's position and occurrence of two consecutive characters on a same level as presented in Figure~\ref{segmented}, makes a challenge for segmentation techniques to work perfectly on such complex text image. 
In such scenario we require explicit segmentation that segments the characters. 
We manually segmented characters from a segmented textline or words as shown in Figure~\ref{char}.
Through empirical methods it becomes impossible to correctly segment the characters from words. 
The acquired images were taken in presence of different illumination which is impacted by surrounding environment. 

\begin{figure}[H]
\centering
\includegraphics[scale=0.9]{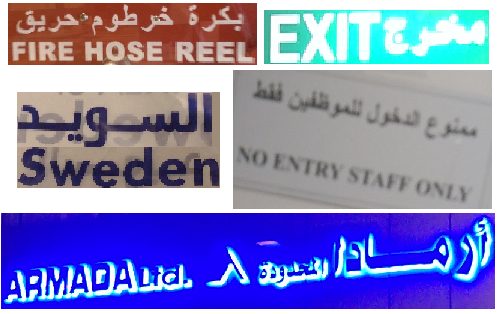}
\caption{\bf Captured images with blur and other impediments}
\label {blur}
\end{figure}

\begin{figure}[H]
\centering
\includegraphics[scale=0.8]{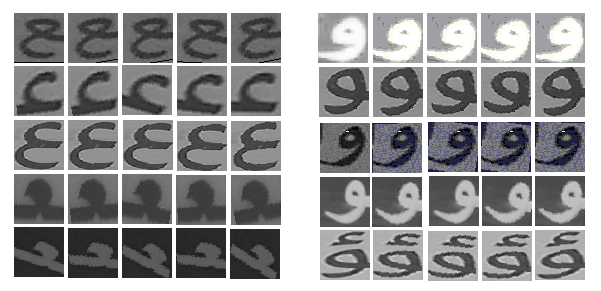}
\caption{\bf Various representations of character "aain" and "wao" with five orientations}
\label{samplei}
\end{figure}

In such situation an impediment is been associated with captured images, such impediment may blur the visibility of a text, such images are represented in Figure~\ref{blur}.
For the purpose to recognize text correctly there is a need to correctly segment text image and remove noise so that classifier may correctly classify the features, learn and recognize the text.
We identified $27$ classes in Arabic script. 
Every class is represented by $20$ images in trainset as depicted in Figure~\ref{samplei}. 
We consider five various orientations of each character. 
As summarized in Table~\ref{tab1}, we have identified $100$ characters representation for each class.
In testset, each class is represented by $5$ variant positions.
After having oriented images we identified $20$ samples for each class.

\begin{table}
  \centering
  \begin{tabular}{|P{2.5cm}|P{1.5cm}|}
    \hline
 
 \textbf{Number of characters}  & $2700$ \\
 \hline
\textbf{Classes}  & $27$ \\
\hline
\textbf{Sample per class with oriented images} & $100$\\
 \hline
 \textbf{Training set} & $2450$\\
 \hline
 \textbf{Test set} & $250$\\
 \hline
\end{tabular}
\newline\newline
\caption{Dataset Statistics}
\label{tab1}
\end{table}

The scene text image is manually segmented into different text lines for example we segmented scene text image into $6$ text lines as represented in Figure~\ref{segmented}. 
	
We performed experiments on various parameters like changing filter size, and a learning rate.
We reported best accuracy when the filter size is $3$ x $3$ and learning rate is $0.005$.

\subsection{Experiments}
Experiments have been performed on limited number of dataset. 
There is no publicly available dataset for Arabic scene text recognition.
So, we are preparing comprehensive Arabic scene text dataset, but currently, we performed experiments on subset of our collected data.
We conducted experiments according to the parameters mentioned in Table~\ref{tab2} .
\begin{table}
  \centering
  \begin{tabular}{|P{1.0cm}|P{1.0cm}|P{1.0cm}|P{1.0cm}|}
    \hline
    \textbf{Filter Size}  & \textbf{Stride}   &  \textbf{Learning Rate}  & \textbf{Error Rate (\%)} \\ \hline
    3 x 3 & 1 & 0.005 & 14.57\\ \hline
    3 x 3 & 1 & 0.5 & 20.93\\ \hline
    3 x 3 & 2 & 0.005 & 18.24\\ \hline
    3 x 3 & 2 & 0.5 & 25.59\\ \hline
    5 x 5 & 1 & 0.005 & 19.75\\ \hline
    5 x 5 & 1 & 0.5 & 29.01\\ \hline
    5 x 5 & 2 & 0.005 & 22.20\\\hline
    5 x 5 & 2 & 0.5 & 33.97\\\hline
  \end{tabular}
  \newline\newline
  \caption{Experimental parameters with error rates}\label{tab2}
\end{table}

Training and testing samples have distributed on the underlaying $27$ identified classes.
Every segmented character is rescaled ($50$ x $50$) and oriented into five different angles.
We performed training on $2450$ character images while trained network is evaluated on $250$ images.
The CovNets has been implemented with $2$ convolutional layers followed by $1$ fully connected layer. 
Both convolutional layers uses $5$ x $5$ convolutions with stride value $2$.
The error rate was reported on $27.01$\%.
In another setting, we introduce max-pooling after each convolutional layer and add an extra fully connected layer with stride value $1$.
The filter size is $5$ x $5$ whereas, learning rate is empirically experimented.
In this way $19.57$\% error rate is measured.

The best accuracy is reported on $3$ x $3$ filter size instead of $5$ x $5$.
The reason to choose minimum filter size is to capture more details about the character image, as Arabic characters also appears with diacritics.
Moreover, we may have more details in pixels about the image.
As learning rate is empirically selected, $14.57$\% error rate is been delineated on $0.005$ learning rate.
The detail about our performed experiments with observed error rate have summarized in Table~\ref{tab2}

The ConvNets are suitable for instance learning tasks rather than sequence learning. We can not learn context from ConvNets rather may extract detailed features of a given pattern.
The feature's detail scrutinized the given pattern at pixel level by variant filter size. 

\begin{figure}[H]
\centering
\includegraphics[scale=0.8]{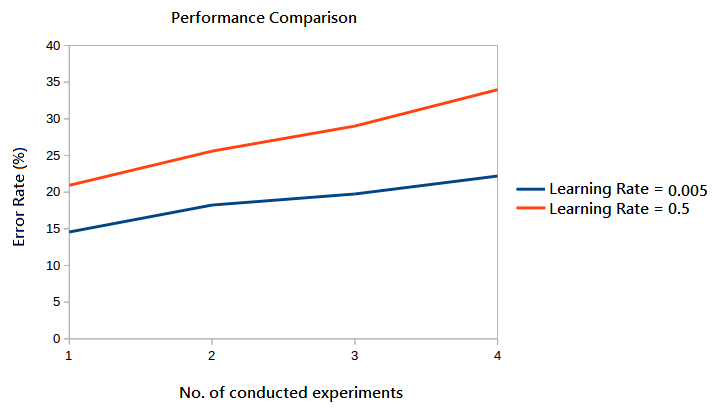}
\caption{\bf ConvNets performance comparison with $3$ x $3$ and $5$ x $5$ filter sizes by keeping learning rate as $0.5$ and $0.005$}
\label {graph}
\end{figure}

As mentioned before that we have evaluated ConvNets on small subset of Arabic scene text images and received encouraging results.
Although there is not any publicly available Arabic scene text dataset but we have investigated ConvNets on subset of our collected data which is been in a process of collection and preparation for Arabic scene text research tasks.
We extracted few variations of $27$ identified Arabic characters.
The best result were reported when filter size was $3$ x $3$ as can be observed in Figure~\ref{graph}.
It is believed and have been noticed from our performed experiments that if filter size is minimum then it may covers more feature which is suitable for languages represented in cursive scripts.

\subsection{Comparison with various feature extraction approaches}
The drawback of ConvNets is that it guaranteed higher accuracy on large dataset.
Most of reported work on cursive scene text recognition obtained good accuracy on huge data. 
As Arabic scene text recognition passing through it's infancy stage, therefore state of the art techniques yet to apply. 
But scene text work on other cursive scripts are available.
We have summarized recent work based on feature extraction approach on various cursive scene texts in Table~\ref{tab3}.

\begin{table}
  \centering
  \begin{tabular}{|P{1.0cm}|P{1.0cm}|P{1.0cm}|P{1.0cm}|}
    \hline
    \textbf{Study}  & \textbf{Script}   &  \textbf{Feature extraction approach}  & \textbf{Error Rate} \\ \hline
    Ren et al~\cite{c1} & Chinese & ConvNets-7 & 0.24\\ \hline
    Ren et al~\cite{c1} & Chinese & ConvNets-9 & 0.31\\ \hline
    Ǵomez et al~\cite{c2}&Multilingual&ConvNets and K-mens &   0.029\\ \hline
    Tounsi et al~\cite{c3}& Arabic & SIFT & 0.24\\ \hline
    Zheng et al~\cite{c4} & Chinese, Japanese, Korean & SIFT & 0.059\\ \hline
   \textbf{Proposed}  & \textbf{Arabic} & \textbf{ConvNets} & \textbf{0.15}\\ \hline
   
  \end{tabular}
  \newline\newline
  \caption{Performance Comparison of cursive scripts scene data with our proposed method}\label{tab3}
\end{table}

As observed from Table~\ref{tab3}, other than our proposed work one more work on Arabic script is proposed in recent years. 
They used scale invariant feature extraction technique. 
Our experiments represented good result in comparison to ~\cite{c3}.
We assumed here that ConvNets extracts more detailed features through its strong layers mechanism whereas, scale invariant feature considered robust but not handling features through layers by which we may get more precise detail of the image in question.

\section{Conclusion}

The ConvNets is suitable to learn patterns of visual images.
The ability to learn without considering the context make it as instance learner.
The potential of investigating the image at pixels level and pool them together on the basis of maximum value make ConvNets a unique deep learning approach.
Such approach is more appropriate in cursive scripts where to extract features is a real challenge.
As the Arabic script has numerous challenges associated like variant shape of characters with respect to positions. 
There is no space in two words which make it harder to segment them with automated tools.
Therefore, by keeping in view these limitations in Arabic script, we used explicit segmentation and feature extraction approaches that may guide us to desired accuracy.
We evaluated the ConvNets deep learning approach on intrinsic Arabic script and report invigorating results.
The experimental results indicates that the ConvNets can improve accuracy on large and variant dataset hence to get better performance on captured Arabic scene text pattern.

\section*{Acknowledgment}
The authors would like to thank Ministry of Education Malaysia and Universiti Teknologi Malaysia for funding this research project through a research Grant (4F801).

% if have a single appendix:
%\appendix[Proof of the Zonklar Equations]
% or
%\appendix  % for no appendix heading
% do not use \section anymore after \appendix, only \section*
% is possibly needed

% use appendices with more than one appendix
% then use \section to start each appendix
% you must declare a \section before using any
% \subsection or using \label (\appendices by itself
% starts a section numbered zero.)
%

% use section* for acknowledgement

% Can use something like this to put references on a page
% by themselves when using endfloat and the captionsoff option.
\ifCLASSOPTIONcaptionsoff
  \newpage
\fi

% trigger a \newpage just before the given reference
% number - used to balance the columns on the last page
% adjust value as needed - may need to be readjusted if
% the document is modified later
%\IEEEtriggeratref{8}
% The "triggered" command can be changed if desired:
%\IEEEtriggercmd{\enlargethispage{-5in}}

% references section

% can use a bibliography generated by BibTeX as a .bbl file
% BibTeX documentation can be easily obtained at:
% http://www.ctan.org/tex-archive/biblio/bibtex/contrib/doc/
% The IEEEtran BibTeX style support page is at:
% http://www.michaelshell.org/tex/ieeetran/bibtex/
%\bibliographystyle{IEEEtran}
% argument is your BibTeX string definitions and bibliography database(s)
%\bibliography{IEEEabrv,../bib/paper}
%
% <OR> manually copy in the resultant .bbl file
% set second argument of \begin to the number of references
% (used to reserve space for the reference number labels box)

% biography section
% 
% If you have an EPS/PDF photo (graphicx package needed) extra braces are
% needed around the contents of the optional argument to biography to prevent
% the LaTeX parser from getting confused when it sees the complicated
% \includegraphics command within an optional argument. (You could create
% your own custom macro containing the \includegraphics command to make things
% simpler here.)
%\begin{biography}[{\includegraphics[width=1in,height=1.25in,clip,keepaspectratio]{mshell}}]{Michael Shell}
% or if you just want to reserve a space for a photo:

\begin{IEEEbiography}[{\includegraphics[width=1in,height=1.25in,clip,keepaspectratio]{picture}}]{John Doe}
\blindtext
\end{IEEEbiography}

% You can push biographies down or up by placing
% a \vfill before or after them. The appropriate
% use of \vfill depends on what kind of text is
% on the last page and whether or not the columns
% are being equalized.

%\vfill

% Can be used to pull up biographies so that the bottom of the last one
% is flush with the other column.
%\enlargethispage{-5in}

% that's all folks
\end{document}